\documentclass[11pt]{article}
\usepackage{coling2020}
\usepackage{times}
\usepackage{url}
\usepackage{latexsym}
\usepackage{graphicx}
\usepackage{linguex}
\usepackage{stackengine}

\setstackgap{L}{.5\normalbaselineskip}
\stackMath
\usepackage{xspace}
\usepackage[hidelinks]{hyperref}
\usepackage{booktabs}
\usepackage{wrapfig}
\usepackage{microtype}
\usepackage{tabularx}
\usepackage{amssymb}
\usepackage{amsmath}
\usepackage{etex}
\usepackage{caption}
\usepackage{calc}
\usepackage{multirow}
\usepackage{xspace}
\usepackage{xcolor}

\newcommand{\emotionname}[1]{\textit{#1}}
\newcommand{\fear}{\emotionname{fear}\xspace}
\newcommand{\joy}{\emotionname{joy}\xspace}
\newcommand{\anger}{\emotionname{anger}\xspace}
\newcommand{\trust}{\emotionname{trust}\xspace}

\newcommand{\surprise}{\emotionname{surprise}\xspace}
\newcommand{\sadness}{\emotionname{sadness}\xspace}
\newcommand{\anticipation}{\emotionname{anticipation}\xspace}
\newcommand{\disgust}{\emotionname{disgust}\xspace}

\newcommand{\noemotion}{\emotionname{no emotion}\xspace}
\newcommand{\avg}{$\varnothing$}
\newcommand{\F}{$\textrm{F}_1$\xspace}

\newcommand{\dsET}{\textit{ElectoralTweets}\xspace}
\newcommand{\dsGNE}{\textit{GoodNewsEveryone}\xspace}
\newcommand{\dsREMAN}{\textit{REMAN}\xspace}
\newcommand{\dsECA}{\textit{Emotion Cause Analysis}\xspace}
\newcommand{\dsES}{\textit{Emotion-Stimulus}\xspace}

\newcommand{\surround}[1]{$\big[$#1$\big]$}
\newcommand{\stack}[2]{$\stackunder{\text{#1}}{\text{\tiny\sf #2}}$}
\newcommand{\target}[1]{\surround{\textcolor{red!70!black}{\stack{#1}{\textsc{target}}}}}
\newcommand{\cue}[1]{\surround{\textcolor{brown!70!black}{\stack{#1}{\textsc{cue}}}}}
\newcommand{\experiencer}[1]{\surround{\textcolor{blue!70!black}{\stack{#1}{\textsc{experiencer}}}}}
\newcommand{\stimulus}[1]{\surround{\textcolor{green!40!black}{\stack{#1}{\textsc{stimulus}}}}}

\newlength\blindedwidth%
\newlength\widthofx%
\newlength\halfblindedwidth%
\newcommand{\blinded}[1]{%
  \setlength{\blindedwidth}{\widthof{#1}}%
  \setlength{\widthofx}{\widthof{X}}%
  \setlength{\halfblindedwidth}{\dimexpr ((\blindedwidth-\widthofx)/2) \relax}%
  \hspace{\halfblindedwidth}X\hspace{\halfblindedwidth}%
}

\newcommand{\plM}{\phantom{$\lfloor$}}
\newcommand{\prM}{\phantom{$\rceil$}}
\newcommand{\lM}{$\lfloor$}
\newcommand{\rM}{$\rceil$}

\colingfinalcopy

\title{Experiencers, Stimuli, or Targets:\\ Which Semantic Roles
  Enable Machine Learning to Infer the Emotions?}

\author{Laura Oberl{\"a}nder, Kevin Reich, \and Roman Klinger \\
  Institut f{\"u}r Maschinelle Sprachverarbeitung, University of Stuttgart, Germany \\
  \texttt{\{firstname.lastname\}@ims.uni-stuttgart.de}\\
}

\date{}

\begin{document}
\maketitle
\begin{abstract}
  Emotion recognition is predominantly formulated as text classification in
  which textual units are assigned to an emotion from a predefined inventory
  (e.g., fear, joy, anger, disgust, sadness, surprise, trust, anticipation).
  More recently, semantic role labeling approaches have been developed to
  extract structures from the text to answer questions like: ``who is
  described to feel the emotion?'' (experiencer), ``what causes this
  emotion?'' (stimulus), and at
  which entity is it directed?'' (target). Though it has been shown that
  jointly modeling stimulus and emotion category
  prediction is beneficial for both subtasks, it remains unclear which of
  these semantic roles enables a classifier to infer the emotion. Is it the
  experiencer, because the identity of a person is biased towards a
  particular emotion ($X$ is always happy)? Is it a particular target
  (everybody loves $X$) or a stimulus (doing $X$ makes everybody sad)? We
  answer these questions by training emotion classification models on five
  available datasets annotated with at least one semantic role by masking the
  fillers of these roles in the text in a controlled manner and find that
  across multiple corpora, stimuli and targets carry emotion information,
  while the experiencer might be considered a confounder.  Further, we
  analyze if informing the model about the position of the role improves the
  classification decision. Particularly on literature corpora we find that
  the role information improves the emotion classification.
\end{abstract}

\section{Introduction}
\blfootnote{
  \hspace{-0.65cm}  
  This work is licensed under a Creative Commons 
  Attribution 4.0 International License.
  License details:
  \url{http://creativecommons.org/licenses/by/4.0/}.
}
Emotion analysis is now an established research area which finds application in
a variety of different fields, including social media analysis
\cite[i.a.]{Purver2012,Wang2012b,Mohammad2017,Ying2019}, opinion mining
\cite[i.a.]{Choi2006}, and computational literary studies
\cite[i.a.]{Ovesdotter2005,Kimfanfic2019,Haider2020,Zehe2020}. The most
prominent task in emotion analysis is emotion categorization, where text
receives assignments from a predefined emotion inventory, such as the
fundamental emotions of \fear, \anger, \joy, \anticipation, \trust, \surprise,
\disgust, and \sadness which follow theories by \newcite{Ekman1999} or
\newcite{Plutchik2001}. Other tasks include the recognition of affect values,
namely valence or arousal \cite{Posner2005} or analyses of event appraisal
\cite{Hofmann2020,Scherer2005}.

More recently, categorization (or regression) tasks have been complemented by
more fine-grained analyses, namely emotion stimulus detection and role
labeling, to detect which words denote the experiencer of an emotion, the
emotion cue description, or the target of an emotion. These efforts lead to
computational approaches of detecting stimulus clauses
\cite{Xia2019,Wei2020,Gao2017} and emotion role labeling and sequence labeling
\cite{Mohammad2014,Bostan2020,Kim2018,Ghazi2015,Zehe2020}, with different
advantages and disadvantages we discuss in \newcite{Oberlaender2020}.

Further, this work led to a rich set of corpora with annotations of
different subsets of roles. An example of a sentence annotated with
semantic role labels for emotion is ``\experiencer{John}
\cue{hates} \target{cars} because they \stimulus{pollute the
  environment}.'' A number of English-language resources are
available: \newcite{Ghazi2015} manually construct a dataset following
FrameNet's emotion predicate and annotate the stimulus as its core
argument.  \newcite{Mohammad2014} annotate Tweets for emotion cue
phrases, emotion targets, and the emotion stimulus. In our previous
work \cite{Bostan2020} we publish news headlines annotated with the
roles of emotion experiencer, cue, target, and
stimulus. \newcite{Kim2018} annotate sentence triples taken from
literature for the same roles.  A popular benchmark for emotion
stimulus detection is the Mandarin corpus by \newcite{Gui2016}.
\newcite{Gao2017} annotate English and Mandarin texts in a comparable
way on the clause level (\dsECA, ECA).

In this paper, we utilize role annotations to understand their
influence on emotion classification. We evaluate which of the roles'
contents enable an emotion classifier to infer the emotions. It is
reasonable to assume that the roles' content carries different kinds
of information regarding the emotion: One particular experiencer
present in a corpus might always feel the same emotion; hence, be
prone to a bias the model could pick up on. The target or stimulus
might be independent of the experiencer and be sufficient to infer the
emotion.  The presence of a target might limit the set of emotions
that can be triggered.  Finally, as some of the corpora contain cue
annotations, we assume that these are the most helpful to decide on
the expressed emotion, as they typically have explicit references
towards concrete emotion names.

\section{Experimental Setting}
\label{sec:experimentalsetting}
In the following, we describe our experiments to understand which of the
datasets' annotated roles contribute to the emotion classification performance.

\begin{table}
  \centering
  \small
  \setlength{\tabcolsep}{5pt}
  \begin{tabular}{lrrrrrrrrrr}
    \toprule
    & \multicolumn{2}{c}{Whole Instance} & \multicolumn{2}{c}{Stimulus} & \multicolumn{2}{c}{Cue} & \multicolumn{2}{c}{Target} & \multicolumn{2}{c}{Exp.} \\
    \cmidrule(r){1-1}\cmidrule(lr){2-3}\cmidrule(lr){4-5}\cmidrule(lr){6-7}\cmidrule(lr){8-9}\cmidrule(lr){10-11}
    Dataset &  \# & \avg len & \# & \avg len & \# & \avg len & \# & \avg len & \# & \avg len \\
    \cmidrule(r){1-1}\cmidrule(lr){2-3}\cmidrule(lr){4-5}\cmidrule(lr){6-7}\cmidrule(lr){8-9}\cmidrule(lr){10-11}
    \dsES, \newcite{Ghazi2015} & 2414 & 20.60 & 820 & 7.29 & --- & --- & --- & --- & --- & ---\\
    \dsET, \newcite{Mohammad2014} & 4056 & 19.14 & 2427 & 6.25 & 2930 & 5.08 & 2824 & 1.71 & 29 & 1.76 \\
    \dsGNE, \newcite{Bostan2020} & 5000 & 13.00 & 4798 & 7.29 & 4736 & 1.60 & 4474 & 4.86 & 3458 & 2.03\\
    \dsREMAN, \newcite{Kim2018} & 1720 & 72.03 & 609 & 9.33 & 1720 & 3.82 & 706 & 5.35 & 1050 & 2.04\\
    \dsECA, \newcite{Gao2017} & 2558 & 62.24 & 2485 & 9.52 & --- & --- & --- & --- & --- & --- \\
    \bottomrule
\end{tabular} \caption{Datasets with annotations of
  roles. \# refers to the number of total instances. \avg{}len shows
  the average length of each role filler in each dataset in the number of tokens.}
\label{tab:datasets}
\end{table}

\paragraph{Datasets.}
We base our experiments on five available datasets that are annotated
for at least one of the roles of an experiencer, stimulus, target, or
cue. The dataset by \newcite{Ghazi2015} is one of the earliest we are
aware of that contains stimulus annotations. They annotate based on
FrameNet's \emph{emotion-directed} frames that have a stimulus argument
in the data (we refer to their
corpus as \dsES, ES). Similarly early work is the Twitter corpus by
\newcite{Mohammad2014} (\dsET, ET). They also follow the emotion frame
semantics definition but use data concerning the 2012 U.S\@.
election. Therefore, their resource may be considered more diverse in
language but more consistent in its domain than ES.
More recently, \newcite{Bostan2020} published an annotation of news
headlines (\dsGNE, GNE). While they do not limit their corpus on a
domain, they use a comparably narrow time window to retrieve the data
and sample according to the inclusion of emotion words and popularity
on social media. \newcite[\dsREMAN]{Kim2018} and \newcite[\dsECA,
ECA]{Gao2017} use literature data, which might be considered the most
challenging for emotion analysis (for ECA, we use the English subset
only).

As Table~\ref{tab:datasets} shows, the literature data (REMAN, ECA)
has the longest instances and also the longest stimulus
annotations. The other resources have less than one third of their length in
tokens, with GNE being the shortest. However, the overall annotation
length does not differ dramatically. Cue, target, and experiencer
annotations are only available in three out of five corpora (ET,
REMAN, and GNE)\footnote{For ET, 90\% of the annotated experiencers
  are the authors of the tweets without corresponding span
  annotation.}.

\paragraph{Model Configuration.}
Our goal is to analyze the importance of different roles for the emotion
classification. We use two different models, namely a bidirectional long
short-term memory network \cite{Hochreiter1997} with pretrained 300-dimensional
GloVe embeddings\footnote{We use 42B tokens, pretrained on CommonCrawl
\cite{Pennington2014}, \url{https://nlp.stanford.edu/projects/glove/}} and
a transformer-based model, RoBERTa \cite{Liu2019}. Both models take as input
the text sequence and output the emotion class, where the concrete set of
emotion labels depends on the dataset.

The models have different advantages and disadvantages in our experimental
setting. The bi-LSTM with non-contextualized word embeddings might be more
appropriate to be used in our setting in which we manipulate the input token
sequence (see below). The transformer might benefit from the rich
contextualized pretraining, which is particularly relevant given that the
annotated corpora are of comparably limited size (in the context of deep
learning)\footnote{The hyperparameters and details for the models are as
follows. For the bi-LSTM, we set a dropout and recurrent dropout of 0.3 and
optimize with Adam \cite{KingmaB14}, with a base learning rate of 0.0003, L2
regularization, on a batch size of 32, with early stopping with patience of 3,
and initialization with Kaiming \cite{He2015}. We train for up to 100 epochs for
the bi-LSTM model and 10 for the transformer-based model. Both models fine-tune
their input representations during training. The hyperparameters of the model
are optimized for ECA. For the bi-LSTM, we
use AllenNLP \cite{allennlp} and for the transformer the Hugging Face library
\cite{wolf2019} (following the training procedure described by
\newcite{devlin2019}). The code of our project is available at
\url{http://www.ims.uni-stuttgart.de/data/emotion-classification-roles}.}.

\paragraph{Setting and Hypotheses.}
\begin{wraptable}[13]{R}{0.60\textwidth}
  \vspace{-0.5\baselineskip}
  \addtolength{\tabcolsep}{-4pt}
  \centering\small
  \begin{tabular}{ll}
    \toprule
    Setting & Model Input \\
    \cmidrule(r){1-1}\cmidrule(l){2-2}
    As-Is & \plM{}John\prM{} hates \plM{}cars\prM{} because they \plM{}pollute the environment\prM{} \\
    Only Stim. &  \plM{}\blinded{John}\prM{} \blinded{hates} \plM{}\blinded{cars}\prM{} \blinded{because} \blinded{they} \plM{}pollute the environment\prM{} \\
    Only Exp. & \plM{}John\prM{} \blinded{hates} \plM{}\blinded{cars}\prM{} \blinded{because} \blinded{they} \plM{}\blinded{pollute} \blinded{the} \blinded{environment}\prM{} \\
    Only Tar. &  \plM{}\blinded{John}\prM{} \blinded{hates} \plM{}cars\prM{} \blinded{because} \blinded{they} \plM{}\blinded{pollute} \blinded{the} \blinded{environment}\prM{} \\
    Without Stim. & \plM{}John\prM{} hates \plM{}cars\prM{} because they \plM{}\blinded{pollute} \blinded{the} \blinded{environment}\prM{} \\
    Without Exp. & \plM{}\blinded{John}\prM{} hates \plM{}cars\prM{} because they \plM{}pollute the environment\prM{} \\
    Without Tar. & \plM{}John\prM{} hates \plM{}\blinded{cars}\prM{} because they \plM{}pollute the environment\prM{} \\
    Pos. Stim. & \plM{}John\prM{} hates \plM{}cars\prM{} because they \lM{}pollute the environment\rM{} \\
    Pos. Exp. & \lM{}John\rM{} hates \plM{}cars\prM{} because they \plM{}pollute the environment\prM{} \\
    Pos. Tar. & \plM{}John\prM{} hates \lM{}cars\rM{} because they \plM{}pollute the environment\prM{} \\
    \bottomrule
  \end{tabular}
  \caption{Illustration of the experimental settings. X, \lM{}, \rM{} denote 
    special tokens added to the input according to each setting.}
  \label{tab:settings}
\end{wraptable}

We apply these models in several settings (illustrated in
Table~\ref{tab:settings}), which differ in the availability of information from
the roles, namely (1), \textit{As-Is}: This is the standard setting: The
classifier has access to the whole text. (2), \textit{Without} the text of the
particular roles. (3), \textit{Only} with the text of a particular role,
masking the text that does not belong to it. Finally, (4), we keep the
information available as is, but besides inform the model about the
\textit{Position} of the role. The latter is realized by adding positional
indicators, inspired by \newcite{Kim2019} who showed the use of positional
indicators for emotion relation classification\footnote{We experimented with
adding two channels in the input embeddings which mark the tokens outside a
role annotation with a 1 in one channel and the tokens which belong to the role
annotation with a 1 in a second channel. The results were inferior to using
positional indicators.}.

For roles that carry information relevant for emotion classification, we expect
the \textit{Without} setting to show a drop in performance compared to the
\textit{As-Is} setting. In such cases, the \textit{Only} setting might show
comparable performance, and the \textit{Position} setting would show further
improvements. When the role is a confounder, the performance in the
\textit{Without} setting is expected to be increased over the \textit{As-Is}
setting.

The label set depends on each of the datasets. For ES, we use the
emotion labels \anger, \disgust, \fear, \joy, \noemotion, \sadness,
and \surprise; for ECA, we use \anger, \sadness, \disgust, \joy,
\fear, \surprise, and \noemotion.  For GNE and ET, we merge the
categories according to the rules described for ET by
\newcite{Bostan2018} and keep the primary emotions described in
Plutchik's wheel.  For REMAN, we group similarly and keep \anger,
\disgust, \fear, \joy, \anticipation, \surprise, \sadness, \trust, and
\noemotion. ECA has a low number of instances annotated with multiple
labels, which we ignore to keep all tasks as single-label
classification. REMAN has emotion annotations only for the middle
sentence in each triple. Thus we include only these middle segments in
our experiments.

The results are based on a random split of each dataset into train, validation,
and test (0.8, 0.1, 0.1). We report macro-averages across 10 runs for the
bi-LSTM and 5 runs for RoBERTa.

\section{Results}
\begin{table}[t]
\small
\centering
\newcommand{\sep}{\cmidrule(r{1pt}){1-1}\cmidrule(lr{20pt}){2-2}\cmidrule(l{-3pt}r{25pt}){3-5}\cmidrule(l{-3pt}r{25pt}){6-8}\cmidrule(l{-3pt}r{25pt}){9-11}\cmidrule(l{-3pt}){12-14}}
\newcommand{\as}{\hspace{10mm}\mbox{}}
\setlength{\tabcolsep}{7pt}
\begin{tabular}{ll @{\hskip 30pt} rrr @{\hskip 30pt} rrr @{\hskip 30pt} rrr @{\hskip 30pt} rrr}
\toprule
& & \multicolumn{3}{c}{As-Is\as} & \multicolumn{3}{c}{Without\as} & \multicolumn{3}{c}{Only\as}  & \multicolumn{3}{c}{Position}  \\
\cmidrule(l{-3pt}r{25pt}){3-5}\cmidrule(l{-3pt}r{25pt}){6-8}\cmidrule(l{-3pt}r{25pt}){9-11}\cmidrule(l{-3pt}){12-14}
Dataset & Role & P & R & \F & P & R & \F & P & R & \F & P & R & \F \\
\sep
  ECA & Stimulus &
41 & 39 & 39 &
\textbf{48} & \textbf{48} & \textbf{48} &
30 & 25 & 23 &
\textbf{52} & \textbf{51} & \textbf{51}
  \\
\sep
  ES & Stimulus &
93 & 89 & 90 &
\textbf{94} & 89 & 90 &
65 & 23 & 18 &
\textbf{95} & \textbf{90} & \textbf{92}
\\
\sep
  \multirow{4}{*}{REMAN} & Cue &
 \multirow{4}{*}{47} & \multirow{4}{*}{27} & \multirow{4}{*}{25} &
\textbf{61} & 14 & 8 &
\textbf{53} & 14 & 8 &
42 & 23 & 19
\\
& Stimulus &  & & &
41 & 22 & 19 &
\textbf{91} & 11 & 4 &
44 & 14 & 12
\\
& Experiencer &  & & &
29 & 23 & 19 &
\textbf{60} & 11 & 6 &
32 & 25 & 21
\\
& Target &  & & &
19 & 12 & 9 &
\textbf{57} & 10 & 3 &
31 & 23 & 21 \\
\sep
\multirow{3}{*}{ET} & Cue & \multirow{3}{*}{51} & \multirow{3}{*}{26} & \multirow{3}{*}{25} & \textbf{63} & 23 & 22 & \textbf{79} & 18 & 15 & \textbf{62} & 25 & 23 \\
& Stimulus & & && 50 & 23 & 21 & \textbf{59} & 15 & 11 & \textbf{57} & \textbf{27} & \textbf{27} \\
& Experiencer & & & & \textbf{53} & 26 & 24 & \textbf{80} & 12 & 7 & 48 & 23 & 20 \\
& Target & & & & \textbf{56} & \textbf{27} & \textbf{26} & \textbf{64} & 16 & 14 & \textbf{65} & 24 & 21 \\
\sep
\multirow{4}{*}{GNE} & Cue & \multirow{4}{*}{34} & \multirow{4}{*}{14} & \multirow{4}{*}{12} & \textbf{62} & 13 & 10 & \textbf{93} & 10 & 5 & \textbf{64} & 13 & 10 \\
& Stimulus & & & &\textbf{93} & 10 & 5 & \textbf{85} & 11 & 7 & \textbf{60} & 13 & 9 \\
& Experiencer & & & & \textbf{55} & \textbf{18} & \textbf{15} & \textbf{93} & 10 & 5 & \textbf{63} & \textbf{15} & \textbf{13} \\
& Target & & & & \textbf{86} & 12 & 8 & \textbf{93} & 10 & 5 & \textbf{62} & 14 & 11 \\
\bottomrule
\end{tabular}
\caption{Results of our bi-LSTM based model for emotion
  classification, with access to all tokens (\textit{As-Is}),
  \textit{Only} to the respective role, to all tokens \textit{Without}
  the respective role, and all tokens together with the
  \textit{Position}al indicators of the role added. All \F scores are
  macro averaged, the scores which are higher than in the As-Is
  setting are bold.}
\label{tab:results}
\end{table}

\begin{table}[t]
\small
\centering
\newcommand{\sep}{\cmidrule(r{1pt}){1-1}\cmidrule(lr{20pt}){2-2}\cmidrule(l{-3pt}r{25pt}){3-5}\cmidrule(l{-3pt}r{25pt}){6-8}\cmidrule(l{-3pt}r{25pt}){9-11}\cmidrule(l{-3pt}){12-14}}
\newcommand{\as}{\hspace{10mm}\mbox{}}
\setlength{\tabcolsep}{7pt}
\begin{tabular}{ll @{\hskip 30pt} rrr @{\hskip 30pt} rrr @{\hskip 30pt} rrr @{\hskip 30pt} rrr}
\toprule
& & \multicolumn{3}{c}{As-Is\as} & \multicolumn{3}{c}{Without\as} & \multicolumn{3}{c}{Only\as}  & \multicolumn{3}{c}{Position}  \\
\cmidrule(l{-3pt}r{25pt}){3-5}\cmidrule(l{-3pt}r{25pt}){6-8}\cmidrule(l{-3pt}r{25pt}){9-11}\cmidrule(l{-3pt}){12-14}
Dataset & Role & P & R & \F & P & R & \F & P & R & \F & P & R & \F \\
\sep
ECA & Stimulus & 68 & 70 & 68 & 4 & 17 & 7 & 4 & 17 & 7 & \textbf{73} & \textbf{73} & \textbf{73}  \\
\sep
ES & Stimulus & 99 & 98 & 98 & 99 & \textbf{99} & \textbf{99} & 3 & 14 & 5 & 99 & 97 & 98  \\
\sep
\multirow{4}{*}{REMAN} & Cue & \multirow{4}{*}{67} & \multirow{4}{*}{60} & \multirow{4}{*}{66} & 3 & 12 & 5 & 3 & 12 & 5 & \textbf{79} & \textbf{77} & \textbf{78} \\
& Stimulus  &  &  &  & 45 & 54 & 47 & 2 & 11 & 4 & 43 & 47 & 43    \\
& Experiencer & &  &  & 60 & 60 & 56 & 2 & 11 & 4 & 62 & 56 & 56  \\ 
& Target &  &  &  & 46 & 42 & 42 & 2 & 11 & 3 & 44 & 45 & 42   \\
\sep
\multirow{4}{*}{ET} & Cue & \multirow{3}{*}{34} & \multirow{3}{*}{33} & \multirow{3}{*}{34} & 32 & 29 & 30 & 5 & 12 & 7 & 31 & 30 & 30  \\
    & Stimulus &  &  &  & \textbf{37} & 33 & 34 & 9 & 15 & 11 & 33 & 32 & 32  \\
    & Experiencer &  &  &  & 34 & \textbf{34} & 34 & 5 & 12 & 7 & 34 & \textbf{34} & 34  \\
& Target &  &  &  & \textbf{35} & \textbf{34} & 34 & 5 & 12 & 7 & \textbf{35} & 33 & 33 \\
\sep
\multirow{4}{*}{GNE} & Cue & \multirow{4}{*}{32} & \multirow{4}{*}{31} & \multirow{4}{*}{31} & 32 & 27 & 27 & 3 & 10 & 5 & 29 & 28 & 28 \\
& Stimulus &  &  &  & 7 & 11 & 7 & 24 & 23 & 23 & \textbf{35} & \textbf{33} & \textbf{34} \\
& Experiencer &  &  & & 31 & 30 & 30 & 3 & 10 & 5 & \textbf{35} & \textbf{32} & \textbf{33} \\
& Target &  &  &  & 3 & 10 & 5 & 3 & 10 & 5 & \textbf{35} & 31 & \textbf{32}  \\
\bottomrule
\end{tabular}
\caption{Results of our transformer based model (RoBERTa) for emotion classification. }
\label{tab:bert-results}
\end{table}

In the following, we discuss the results of the bi-LSTM model in detail and
then point to differences to those of the transformer-based
approach. Table~\ref{tab:results} shows the results of our experiments
for the bi-LSTM-based model. Intuitively, we would expect the
\textit{As-Is} setting to outperform both the \textit{Without} and
\textit{Only} settings because there is more information available to
the model. Conversely, because information is added in
\textit{Position}, we expect it to outperform the \textit{As-Is}
setting.
As we see in column \textit{As-Is}, the scores for the emotion
classification task differ substantially, even when all available
information is shown to the model. In the \textit{Without} setting, we
see that removing information can sometimes help a model improve its
decision.  For instance, when we mask the labels of the respective
role, we observe a performance increase for the experiencer role in
GNE, which could potentially point to an unwanted bias for particular
experiencers in this corpus. This is also the case for the
stimulus role in ECA and the target role in ET.

As expected, an important role for emotion classification is the
cue. In REMAN, the performance drops the most when the classifier does
not see the cue span and gains the most when only the cue is
available. For all other corpora, the cue role is not as important,
but performance still shows a drop when it is not available
(\textit{Without}).
Similarly, for all datasets except ECA, the performance drops when
the stimulus is not shown. On the other hand, the stimulus alone is
insufficient to infer the emotion with competitive
performance. Noteworthy here is the corpus ES, in which the
performance drop is particularly high.

These results show that the information contained in different roles is of
varying importance and depends on the data's source and domain. In the
setting \textit{Position}, we leave all information accessible to
the model but add positional indicators for the investigated role to
the input for emotion classification. We see improvements in most
cases, except REMAN, for which adding the positional information hurts
the classification for all roles. This result could be because REMAN
has very long annotation spans.  Both ECA and ES show an improvement
for their annotated role (stimulus). For ET, an increase in
performance is shown when additional knowledge about the stimulus
position is given, and for GNE, a slight improvement is shown when the
model is given the experiencer's position information.

Table~\ref{tab:bert-results} shows the results of the
transformer-based model evaluated in the same settings.  As expected,
the model shows performance improvements across all datasets in
comparison to the bi-LSTM model. In the \textit{As-Is} setting, we see
a substantial increase in performance for REMAN. This result can be
explained by the fact that the pretrained large language model has
seen more literary English than the embeddings used as pretrained
input to the bi-LSTM. GNE and ET scores are also improved across the
roles. In the \textit{Without} setting, we do not see the same
patterns as for the bi-LSTM based model; the scores when hiding the
stimulus for ECA, the target for ET, and experiencer for GNE do not
increase over the scores of the \textit{As-Is} setting.

This might have two reasons: On one hand, it is less likely to improve upon already high values when changing the model configuration. On the other hand, and more
interestingly, it might be that the contextualized embeddings
compensate for missing information.  Interestingly for the
\textit{Position} setting, the results are improving on all datasets,
and REMAN gains from the cue's positional indicators. The dataset that
stands out in this setting is ET, for which we see a slight decrease
in performance across all roles available. The \textit{Only} setting
shows that the stimulus captures most of the emotion information for
GNE and ET. The result for GNE is due to the particularly lengthy
stimuli spans that sometimes stretch over the whole instance.

\section{Conclusion and Future Work}
Our experiments show that the importance of semantic roles for emotion
classification differs between datasets and roles: The stimulus and
cue are critical for classification, which correspond to the direct
report of a feeling and the description that triggered an
emotion. This result is shown in the drop in performance when removing
these roles. This information is not redundantly available outside of
these arguments.

It is particularly beneficial for the model's performance to have
access to the position of cues and stimuli. This suggests that the
classifier learns to tackle the problem differently when this
information is available, especially so for ECA and ES -- the cases in
which literature has been annotated and the instances are comparably
long.

The bi-LSTM model indicates that the experiencer role is a confounder
in GNE.  The performance can be increased when the model does not have
access to its content. Similar results are observed for ET, in which
the target role is a confounder. However, these results should be
taken with a grain of salt given that they are not confirmed while
switching to the transformer-based model. The differences in
results between the bi-LSTM and the transformer also motivate further
research, as they suggest that the contextualized representation might
compensate for missing information, and is, therefore, more
robust.

Finally, our results across both models and multiple datasets
indicate that emotion classification approaches indeed benefit from
semantic roles' information by adding the positional
information. Similarly to targeted and aspect-based sentiment
analysis, this motivates future work, in which emotion classification
and role labeling should be modelled jointly. In this case, it can also be interesting
to investigate what happens when the positional
indicators are added to all roles jointly.

\section*{Acknowledgements}
This work was supported by Deutsche
Forschungsgemeinschaft (project SEAT, KL 2869/1-1). We thank Enrica
Troiano and Heike Adel for fruitful discussions and the anonymous
reviewers for helpful comments.

\bibliographystyle{coling}
\bibliography{lit}

\clearpage

\appendix

\section*{Appendix}
\label{sec:appendix}

\subsection*{Qualitative Discussion of Examples}
\label{sec:qualitative}
We analyze a subset of interesting cases from the results section in
the following, to better understand why removing stimuli from ECA
improves the results and further why the same can be observed on ET
for targets.

We show examples for these cases in Table~\ref{tab:examples}. We
observe in instances correctly classified in the \textit{Without}
setting that removing the stimulus makes the classification task
easier by removing potential sources for overfitting: The remaining
tokens contain the explicit cue, even though they are not explicitly
annotated for ECA. For instance, in ``\stimulus{his angry outbreak}
\cue{saddened} \experiencer{me}'', we see that removing the stimulus
which also contains a reference to another emotion, the task of
picking the most dominant emotion from the remaining tokens is more
straight-forward.

This holds similarly for other examples in ECA, in which the stimulus
describes an event that could also be evaluated as scary; however, the
experiencer mentions that he is surprised (``To my surprise'').

\mbox{}\\[\baselineskip]
\begingroup
  \centering
  \footnotesize
  \setlength{\tabcolsep}{2pt}
  \renewcommand*{\arraystretch}{1.3}
  \newcommand{\ms}[1]{\textbf{\textcolor{blue!70!black}{#1}}}
  \newcommand{\sep}{\cmidrule(r){1-1}\cmidrule(rl){2-2}\cmidrule(rl){3-3}\cmidrule(rl){4-4}\cmidrule(rl){5-5}\cmidrule(rl){6-6}\cmidrule(rl){7-7}\cmidrule(rl){8-8}}
  \begin{tabularx}{\linewidth}{lccccccX}
    \toprule
    & \multicolumn{6}{c}{Label} \\
    \cmidrule(l){3-8}
    &      &     & \multicolumn{4}{c}{without} \\
    \cmidrule(l){4-7}
    Dataset & Gold & All & Stim. & Exp. & Cue & Targ. & Text \\
    \sep
      GNE     & \ms{J} & Su & \ms{J} & Su & Su & Su & \experiencer{Djokovic} \cue{happy} \stimulus{to carry on cruising}  \\
      GNE     & \ms{J} & Su & \ms{J} & Su & A & Su &  \experiencer{Trump} \cue{upbeat} \stimulus{on potential for US-Japan trade deal.} \\
    \sep
      ECA & \ms{J} & F & \ms{J} & -- & -- & -- &  \stimulus{``Michie Reetchie''}, said Xavier, and again he burst into laughter that choked further speech. He controlled himself and laid his finger on his vein. \\
      ECA & \ms{Su} & F & \ms{Su} & -- & -- & -- & One morning Pop sent me down to the river to catch some fish for breakfast. To my surprise \stimulus{there was a canoe in the water and there was no one in}. Immediately I jumped into the river and brought the canoe to the side. \\
      ECA & \ms{F} & S & \ms{F} & -- & -- & -- &  I did not answer,                                                 fearing \stimulus{to
                                                 tell him that I had
                                                 been awake watching him} \\
      ECA & \ms{A} & S & \ms{A} & -- & -- & -- & A massy stone and shook the ranks of Troy, as when in anger \stimulus{against long - screaming cranes} a watcher of the field leaps from the ground in swift hand whirling round his head the sling and speeds the stone against them scattering. \\
      ECA & \ms{D} & A & \ms{D} & -- & -- & -- & \stimulus{A year after being fired from his job} he has a lot of resentment towards his former boss. \\
    \sep
      ET & \ms{D} & T & D & T & S & \ms{D} &  Three words to describe the entire \target{\#GOP convention} \cue{Mean and demeaning.} \\
      ET & \ms{A} & D & D & D & D & \ms{A} & \target{\#Republicans} are a joke . \stimulus{Clint Eastwood} is their mascot ! America is in trouble if \cue{these idiots} win ! \#RNC \\
      ET & \ms{J} & T & T & J & T & \ms{J} & \target{Obama Voter} \stimulus{Says Vote for Obama} \cue{YES WE CAN AGAIN !}\\
      ET & \ms{J} & Ant & T & T & T & \ms{J} & \cue{So excited} to vote this upcoming \target{election} \stimulus{finally exercising my right to choose our next president} \#Obama \\
      ET  & \ms{D} & A & A & A & A & \ms{D} & \target{Romney} is gonna put The Onion out of business . \cue{\#TheStench} \\
\sep
      REMAN & \ms{J} & noemo & noemo &\ms{J} & noemo & -- &  And \experiencer{she} returned the quiet but jubilant kiss that he laid upon her lips. \\
    \bottomrule
  \end{tabularx}
  \captionof{table}{\label{tab:examples} Examples in which the prediction is incorrect when the model is applied on the whole instance, but it is correct when the respective role is removed. The correct prediction is marked in bold face. J: Joy, T: Trust, Su: Surprise, Ant: Anticipation, D: Disgust, F: Fear, A: Anger, S: Sadness}
\endgroup

\clearpage

\subsection*{Detailed Results for Additional Positional Information}
\label{sec:results-positional}

We have seen in the results that adding position information of the semantic
roles increases the performance for both datasets which contain examples drawn
from literature. This is particularly interesting for future research on
jointly modelling roles and classification. Therefore, we show details per
emotion class in Table~\ref{tab:esetperemotion} (only for the bi-LSTM model).

We see for the ECA dataset, that when the positional information is made
accessible to the model, the classifier learns better to predict all emotion
classes with a substantial improvement for anger and disgust. Similarly, ES
improves over all emotions with the exception of disgust and sadness.

\begingroup
\vspace{\baselineskip}
\centering
\small
\setlength{\tabcolsep}{15pt}
\begin{tabular}{ll ccc ccc}
  \toprule
  Data & Emotion & \multicolumn{3}{c}{All} & \multicolumn{3}{c}{Stimulus Position}  \\
  \cmidrule(r){1-1} \cmidrule(lr){2-2}\cmidrule(lr){3-5}\cmidrule(l){6-8}
  &          & P & R & \F & P & R & \F \\
 \cmidrule(rl){3-5}\cmidrule(lr){6-8}
  \multirow{6}{*}{ECA}
  & Anger & 15  & 11 & 13 & \textbf{36} & \textbf{44} & \textbf{40}  \\
 & Disgust & 25 & 06 & 09 &  11 & \textbf{11} & \textbf{11}  \\
 & Fear & 56 & 56 & 56 & \textbf{78} & \textbf{70} & \textbf{74} \\
 & Joy & 57 & 58 & 57  & \textbf{65} & 58 & \textbf{61}  \\
 & Sadness & 50 & 67  & 57 & \textbf{57} & \textbf{72} & \textbf{64} \\
 & Surprise & 40 & 38  & 39  & \textbf{63}  & \textbf{53} & \textbf{58}\\
 \cmidrule(lr){2-2}\cmidrule(lr){3-5}\cmidrule(l){6-8}
 & Macro & 40 & 39  & 38 & \textbf{52} & \textbf{51} & \textbf{51}  \\
  \cmidrule(r){1-1} \cmidrule(lr){2-2}\cmidrule(lr){3-5}\cmidrule(l){6-8}
  \multirow{6}{*}{ES}
 & Anger & 90 & 97 & 94  &   \textbf{92} & \textbf{98} & \textbf{95} \\
 & Disgust & 85 & 54 & 67  & \textbf{100} & 45 & 63 \\
 & Fear & 97 & 88 & 93  & 95 & \textbf{95} & \textbf{95}  \\
 & Joy & 93 & 92 & 92 & \textbf{100} & 92 & \textbf{96} \\
 & Sadness & 94 & 99 & 97 &  90 & 96 & 93 \\
 & Shame & 100 & 94 & 97 & 100 & \textbf{100} & \textbf{100} \\
 & Surprise & 91 & 95 & 93 & 88 & \textbf{100} & \textbf{94} \\
\cmidrule(lr){2-2}\cmidrule(lr){3-5}\cmidrule(l){6-8}
 & Macro &  93 & 89 & 90 & \textbf{95} & \textbf{90} & \textbf{91} \\
  \bottomrule
\end{tabular}
\captionof{table}{\label{tab:esetperemotion} Results per emotion for
  ECA and ES with and without positional stimuli information. Bold
  numbers indicate that their value is greater than in the As-Is setting.}
\endgroup

\clearpage

\subsection*{Analysis of Content of Roles}

Table~\ref{tab:token-examples} shows the most frequent tokens marked
as \textit{cue, stimulus, experiencer} or \textit{target} over each
dataset. They differ substantially per dataset and reflect well the
respective source. The counts suggest a Zipfian distribution for \dsET
(stimulus and target) and \dsGNE (experiencer, stimulus). This could
explain the results obtained in the \textit{Without} setting by the
bi-LSTM-based model. The most common tokens annotated with the
\textit{target} role in \dsET also show the polarized nature of those
who tweeted about the election.

\begin{figure}[b]
  \centering
  \includegraphics[page=1,width=0.49\textwidth]{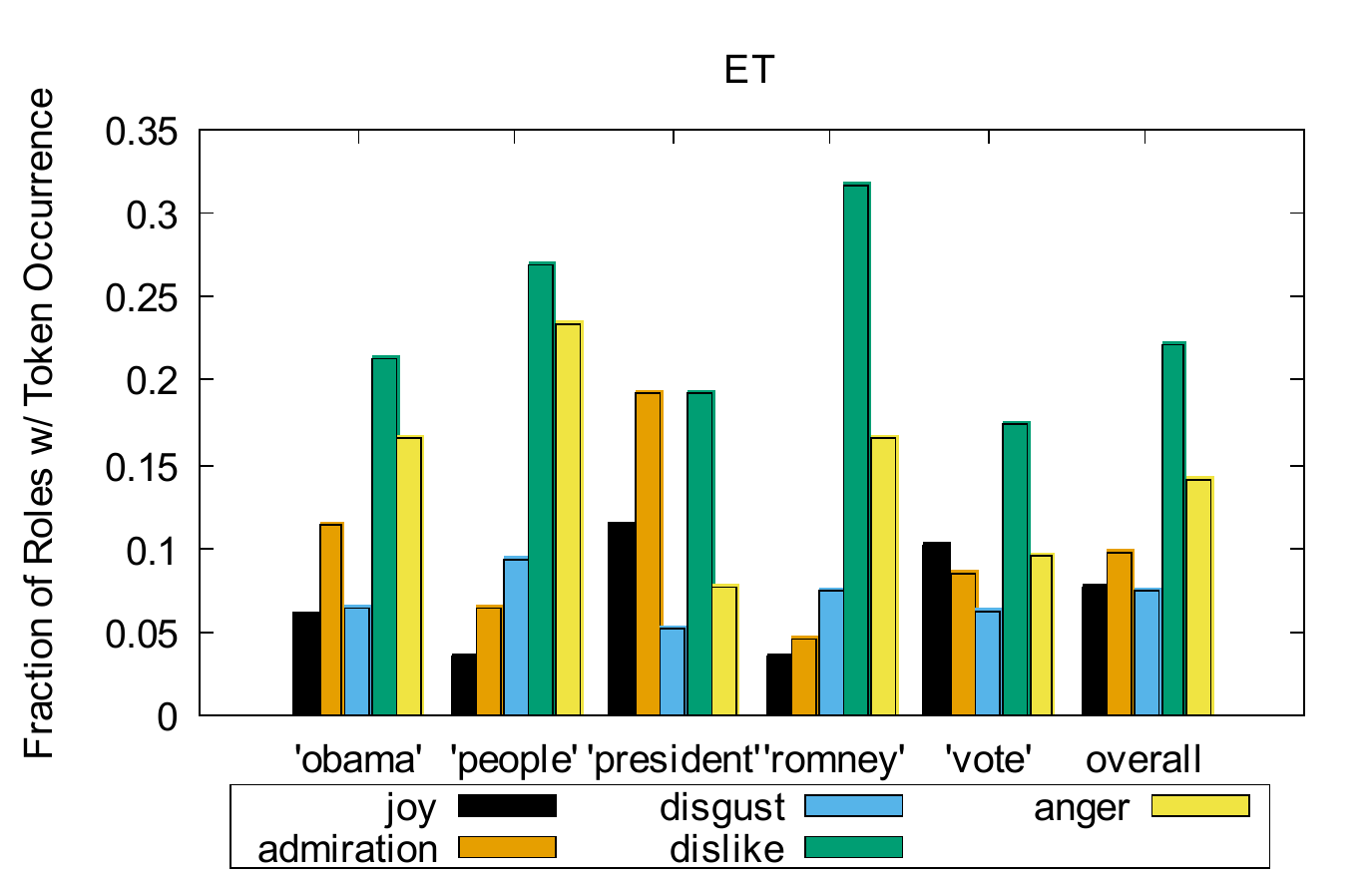}
  \includegraphics[page=2,width=0.49\textwidth]{plot}
  \caption{Emotion distribution of instances containing the respective
    tokens (\% for the top-5 most frequent emotions for each
    dataset). ``overall'' represents the emotion distribution for those
    emotions across all instances.}
    	\label{fig:token}
\end{figure}

Figure~\ref{fig:token} shows the distribution of the most frequent
tokens (across all roles) for the most frequent emotions of ET and
GNE. The plots marked with ``overall'' show the prior distribution of
emotions in the respective dataset. We see that for the emotion
\textit{admiration}, ``president'' stands out. Further we note that
``Romney'' is associated with \textit{dislike} in this corpus.

For GNE we observe that the most frequent tokens are occurring less in
instances annotated with \textit{positive surprise} than overall, and
more in instances annotated with \anger (except for ``Biden'') showing
that these tokens could be biased towards more negative emotions. This
shows a bias of the dataset towards negative emotion when it comes to
the most prominent tokens.\\[1mm]

\begingroup
\centering
\small
\renewcommand*{\arraystretch}{0.9}
\setlength\tabcolsep{1mm}
\begin{tabularx}{\linewidth}{llX}
\toprule
 & Role  & Tokens \\
\cmidrule(r){1-1}\cmidrule(lr){2-2}\cmidrule(l){3-3}
ECA & Stim. & see (80), like (49), man (49), go (43), life (43), father (43), time (42), day (34), came (33), son (32) \\
\cmidrule(r){1-1}\cmidrule(lr){2-2}\cmidrule(l){3-3}
ES & Stim. & see (36), way (12), find (11), left (9), people (9), prospect (8), thought (8), like (8), losing (8), work (7) \\
\cmidrule(r){1-1}\cmidrule(lr){2-2}\cmidrule(l){3-3}
 \multirow{5}{*}{\rotatebox{90}{REMAN}} & Cue  &  love (32), suddenly (31), afraid (15), smile (12), beautiful (11), trust (11), pleasure (10), ugly (7), things (7), wish (6) \\
& Stim. & little (10), another (8), face (8), got (7), lord (7), left (7), great (7), wife (7), men (6), life (6) \\
& Exp. & man (23), woman (12), boy (7), old (7), isabel (6), people (6), god (5), father (5), heart (5), henry (5) \\
& Target & man (22), little (9), things (8), woman (8), see (8), old (7),  god (6), wife (6), another (6), true (5) \\
\cmidrule(r){1-1}\cmidrule(lr){2-2}\cmidrule(l){3-3}
 \multirow{7}{*}{ET} & Cue  & Obama (136), Romney (105), vote (89), like (65), Mitt (56), people (53), get (52), president (50), \par really (49), excited (49) \\
& Stim. & Obama (249), Romney (211), vote (108), Mitt (87), Barack (74), president (66), people (51),\par speech (40), like (40), get (35)  \\
& Exp. & gop, anyone, presidency, clint \\
& Target &  Obama (446), Romney (420), Mitt (146), Barack (112), People (53), president (40), election (20), debate (19), Michelle (19), Clinton (15) \\
\cmidrule(r){1-1}\cmidrule(lr){2-2}\cmidrule(l){3-3}
\multirow{6}{*}{GNE} & Cue  & killed (38), crisis (33), attacks (33), death (26), war (25), arrested (24), racist (24), help (22), new (20), fight (19) \\
 & Stim. & Trump (279), border (68), Mueller (58), back (57), report (56), Iran (57), report (56), war (55),\par people (55), deal (55) \\
& Exp. & Trump (401), Donald (66), man (46), democrats (44), Biden (40), House (37), woman (36), police (35),\par Mueller (34), Sanders (33) \\
& Target & Trump (345), new (94), Mueller (54), House (44), border (43), people (42), democrats (41), deal (36),\par report (36), president (35)  \\
\bottomrule
\end{tabularx}
\captionof{table}{Most frequent 10 tokens with frequencies for each role and dataset.}
\label{tab:token-examples}
\endgroup

\end{document}